\documentclass{article}

\usepackage{arxiv}

\usepackage[utf8]{inputenc} 
\usepackage[T1]{fontenc}    
\usepackage{hyperref}       
\usepackage{url}            
\usepackage{booktabs}       
\usepackage{amsfonts}       
\usepackage{nicefrac}       
\usepackage{microtype}      
\usepackage{graphicx}
\usepackage{amsfonts,amssymb}
\usepackage{algorithm}
\usepackage{algorithmic}
\usepackage{float}
\usepackage{stfloats}
\usepackage{amsmath}
\usepackage{subfigure}
\usepackage{graphicx}
\usepackage{lineno}

\title{Dual Adversarial Semantics-Consistent Network for\\
Generalized Zero-Shot Learning}

\author{
   Jian Ni\textsuperscript{1} 
   \And
   Shanghang Zhang\textsuperscript{2} \\
   University Of Science And Technology Of China\textsuperscript{1,3} \\
   Carnegie Mellon University\textsuperscript{2} \\
   \texttt{nj1@mail.ustc.edu.cn} \\
   \And
   Haiyong Xie\textsuperscript{3} \\
}

\begin{document}
\maketitle

\begin{abstract}
 Generalized zero-shot learning (GZSL) is a challenging class of vision and knowledge transfer problems in which both seen and unseen classes appear during testing. Existing GZSL approaches either suffer from semantic loss and discard discriminative information at the embedding stage, or cannot guarantee the visual-semantic interactions. To address these limitations, we propose the Dual Adversarial Semantics-Consistent Network (DASCN), which learns primal and dual Generative Adversarial Networks (GANs) in a unified framework for GZSL. In particular, the primal GAN learns to synthesize inter-class discriminative and semantics-preserving visual features from both the semantic representations of seen/unseen classes and the ones reconstructed by the dual GAN. The dual GAN enforces the synthetic visual features to represent prior semantic knowledge well via semantics-consistent adversarial learning. To the best of our knowledge, this is the first work that employs a novel dual-GAN mechanism for GZSL. Extensive experiments show that our approach achieves significant improvements over the state-of-the-art approaches.
\end{abstract}

\section{Introduction}

In recent years, tremendous progress has been achieved across a wide range of computer vision and machine learning tasks with the introduction of deep learning. However, conventional deep learning approaches rely on large amounts of labeled data and suffer from performance decay in problems with limited training data. On the one hand, objects in the real world have a long-tailed distribution, and obtaining annotated data is expensive. On the other hand, novel categories of objects arise dynamically in nature, which fundamentally limits the scalability and applicability of supervised learning models for handling this dynamic scenario when labeled examples are not available.

Tackling such restrictions, zero-shot learning (ZSL) has been researched widely recently and recognized as a feasible solution \cite{lampert2013attribute, socher2013zero}. ZSL is a learning paradigm that tries to fulfil the ability to categorize objects from previous unseen classes correctly
\begin{figure}[ht]
\centering
\includegraphics[scale=0.45]{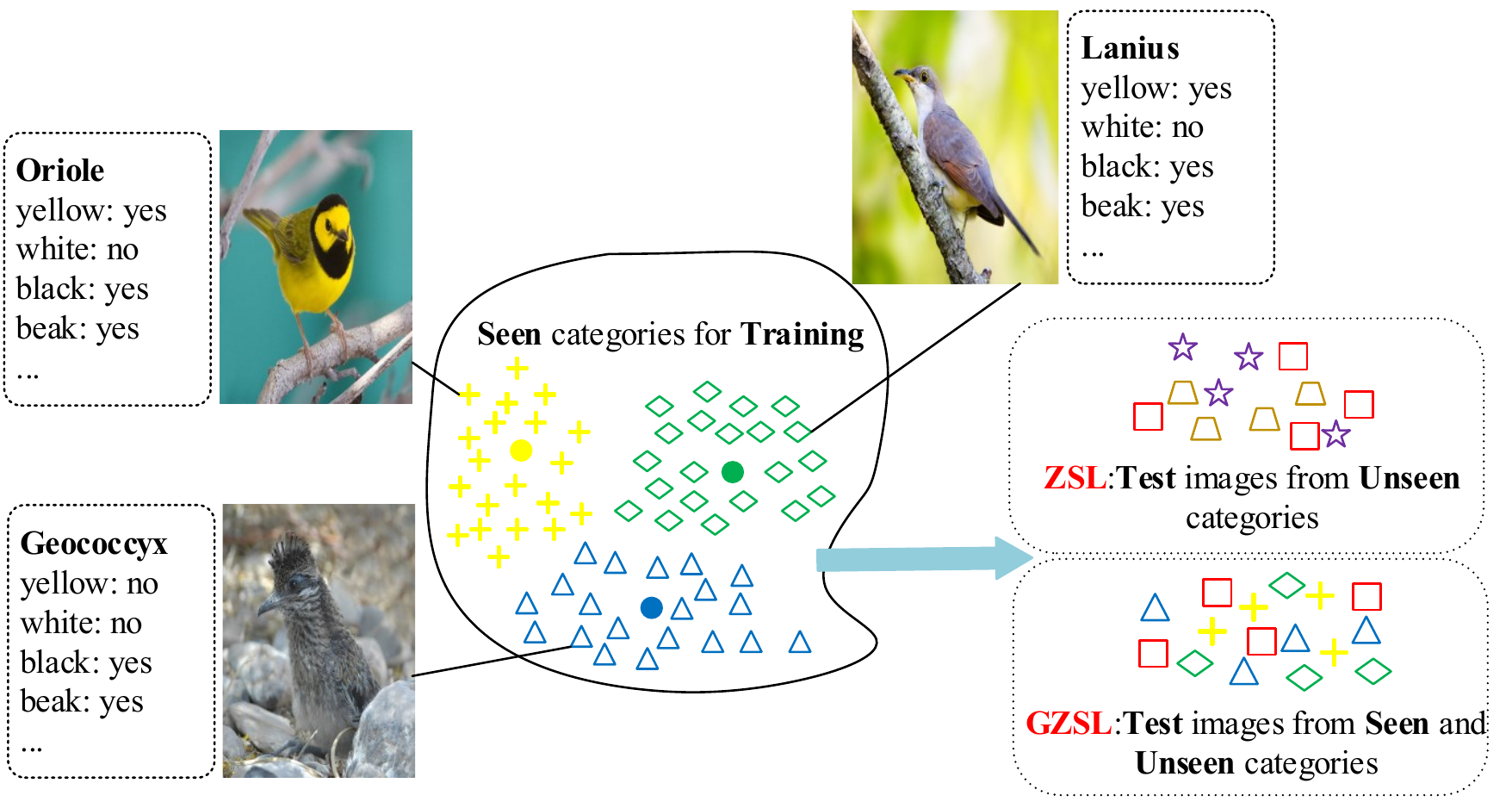}
\label{fig2}
\caption{Problem illustration of zero-shot learning (ZSL) and generalized zero-shot learning (GZSL).}
\end{figure}
without corresponding training samples. However, conventional ZSL models are usually evaluated in a restricted setting where test samples and the search space are limited to the unseen classes only, as shown in Figure 1. To address the shortcomings of ZSL, GZSL has been considered in the literature since it not only learns information that can be transferred to an unseen class but can also generalize to new data from seen classes well.

Typical ZSL approaches consider tasks to be visual-semantic embedding problems \cite{chen2018zero, romera2015embarrassingly, wang2018zero, zhang2017learning}. They try to learn a mapping function from the visual space to the semantic space where all the classes reside, or to a latent intermediate space, so as to transfer knowledge from the seen classes to the unseen classes. However, the ability of these embedding-based ZSL models to transfer semantic knowledge is limited by the semantic loss and the heterogeneity gap \cite{chen2018zero}. Meanwhile, since the ZSL model is only trained with the labeled data from the seen classes, it is highly biased towards predicting the seen classes \cite{chao2016empirical}. Another popular strategy for ZSL is to use generative methods to generate various visual features conditioned on semantic feature vectors \cite{felix2018multi, guo2017synthesizing, long2017zero, xian2018feature, zhu2018generative}, which circumvents the need for labeled samples of unseen classes and boosts the ZSL classification accuracy. Nevertheless, the performance of these methods is limited by either capturing the visual distribution information via only a unidirectional alignment from the class semantics to the visual feature only, or adopting just a single Euclidean distance as the constraint to preserve the semantic information between generated high-level visual features and real semantic features. Recent work has shown that the performance of most ZSL approaches drops significantly in the GZSL setting \cite{xian2018zero}.

In this paper, we propose a Dual Adversarial Semantics-Consistent Network (DASCN) for GZSL based on the Generative Adversarial Networks (GANs) in a dual structure for common visual features generation and corresponding semantic features reconstruction, respectively, as shown in Figure 2. Such bidirectional synthesis procedures boost these two tasks collaboratively by preserving the visual-semantic consistency and make the following two contributions. First, our generative model synthesizes inter-class discrimination visual features via a classification loss constraint, which makes sure that synthetic visual features are discriminative enough among different classes. Second, our model encourages the synthesis of visual features that represent their semantic features well and are of a highly discriminative semantic nature from the perspectives of both form and content. From the form perspective, the semantic reconstruction error between the synthetic semantic features, which are reconstructed by the dual GAN from the pseudo visual features that are generated by the primal GAN, and the real corresponding semantic features is minimized to ensure that the reconstructed semantic features are tightly centered around the real corresponding class semantics. From the content perspective, the pseudo visual features, generated via the primal GAN further exploiting the reconstructed semantic features as input are constrained to be as close as possible to their respective real visual features in the data distribution. Therefore, the method ensures that the reconstructed semantic features are consistent with the relevant real semantic knowledge and avoids semantic loss to a large extent.

The contributions of this paper are summarized as follows: (1) We propose a novel generative dual adversarial architecture for GZSL, which preserves semantics-consistency effectively with a bidirectional alignment and alleviates the issue of semantic loss. To the best of our knowledge, DASCN is the first network to employ the dual-GAN mechanism for GZSL. (2) By combining the classification loss and the semantics-consistency adversarial loss, our model generates high-quality visual features with inter-class
separability and a highly discriminative semantic nature, which is crucial to the generative approaches used in GZSL. (3) Comprehensive experimental results on four benchmark datasets demonstrates the effectiveness of our proposed approach, which consistently outperforms the state-of-the-art GZSL methods consistently.  

\section{Related Work}

\subsection{Zero-Shot Learning}

Some of the early ZSL works make use of the primitive attributes prediction and classification, such as DAP \cite{lampert2009learning}, and IAP \cite{lampert2013attribute}. Recently, the attribute-based classifier has evolved into the embedding-based framework, which now prevails due to its simple and effective paradigm \cite{akata2015evaluation, romera2015embarrassingly, socher2013zero, sung2018learning, zhang2017learning}. The core of such approaches is to learn a projection from visual space to semantic space spanned by class attributes \cite{romera2015embarrassingly, socher2013zero}, or conversely \cite{zhang2017learning}, or jointly learn an appropriate compatibility function between visual and semantic space \cite{akata2015evaluation, sung2018learning}. 

The main disadvantage of the above methods is that the embedding process suffers from semantic loss and the lack of visual training data for unseen classes, thus biasing the prediction towards the seen classes and undermining seriously the performance of models in the GZSL setting. More recently, generative approaches are promising for GZSL setting by generating labeled samples for the seen and unseen classes. \cite{guo2017synthesizing} synthesize samples by approximating the class conditional distribution of the unseen classes based on learning that of the seen classes. \cite{xian2018feature, zhu2018generative} apply GAN to generate visual features conditioned on class descriptions or attributes, which ignore the semantics-consistency constraint and allow the production of synthetic visual features that may be too far from the actual distribution. \cite{felix2018multi} consider minimizing L2 norm between real semantics and reconstructed semantics produced by a pre-trained regressor, which is rather weak and unreliable to preserve high-level semantics via the Euclidean distance. Compared to these approaches, the DASCN learns the semantics effectively via multi-adversarial learning from both the form and content perspectives.

Note that ZSL is also closely related to domain adaptation and image-to-image translation tasks, where all of them assume the transfer between source and target domains. Our approach is motivated by, and is similar in spirit to, recent work on synthesizing samples for GZSL \cite{xian2018feature} and unpaired image-to-image translation \cite{kim2017learning, yi2017dualgan, zhu2017unpaired}. Our model preserves the visual-semantic consistency by employing dual GANs to capture the visual and semantic distributions, respectively. 

\subsection{Generative Adversarial Networks}

As one of the most promising generative models, GANs have achieved a series of impressive results. The idea behind GANs is to learn a generative model to capture an arbitrary data distribution via a max-min training procedure, which consists of a generator and a discriminator that work against each other. DCGAN \cite{radford2015unsupervised} extends GAN by leveraging deep convolution neural networks. InfoGAN \cite{chen2016infogan} maximizes the mutual information between the latent variables and generator distribution. In our work, given stabilizing training behavior and eliminating model collapse as much as possible, we utilize WGANs \cite{gulrajani2017improved} as basic models in a dual structure.

\section{Methodology}

\subsection{Formulation}

We start by formalizing the GZSL task and then introduce the architecture and training objectives of the proposed DASCN. Let $D^{Tr}=\big\{(x,y,a)|x\in \mathcal{X},y\in \mathcal{Y}^s,a\in \mathcal{A}\big\}$, in which $D^{Tr}$ stands for the set of $N^s$ training instances of the seen classes, $x\in \mathcal{X}\subseteq \mathbb{R}^K$ represent {\it K}-dimensional visual features extracted from convolution neural networks, $\mathcal{Y}^s$ denote the corresponding class labels, and $a\in \mathcal{A}\subseteq \mathbb{R}^L$ denote semantic features, e.g. attributes of seen classes. In addition, we have a disjoint class label set $\mathcal{U}=\big\{(y,a)|y\in \mathcal{Y}^u,a\in \mathcal{A}\big\}$ of unseen classes, where visual features are missing. Given $D^{Tr}$ and $\mathcal{U}$, in GZSL, we learn a prediction: $\mathcal{X}\to\mathcal{Y}^s \cup \mathcal{Y}^u$. Note that our method is of the inductive school where model has no access to neither visual nor semantic information of unseen classes in the training phase.

\begin{figure*}[ht]
\centering
\includegraphics[scale=0.47]{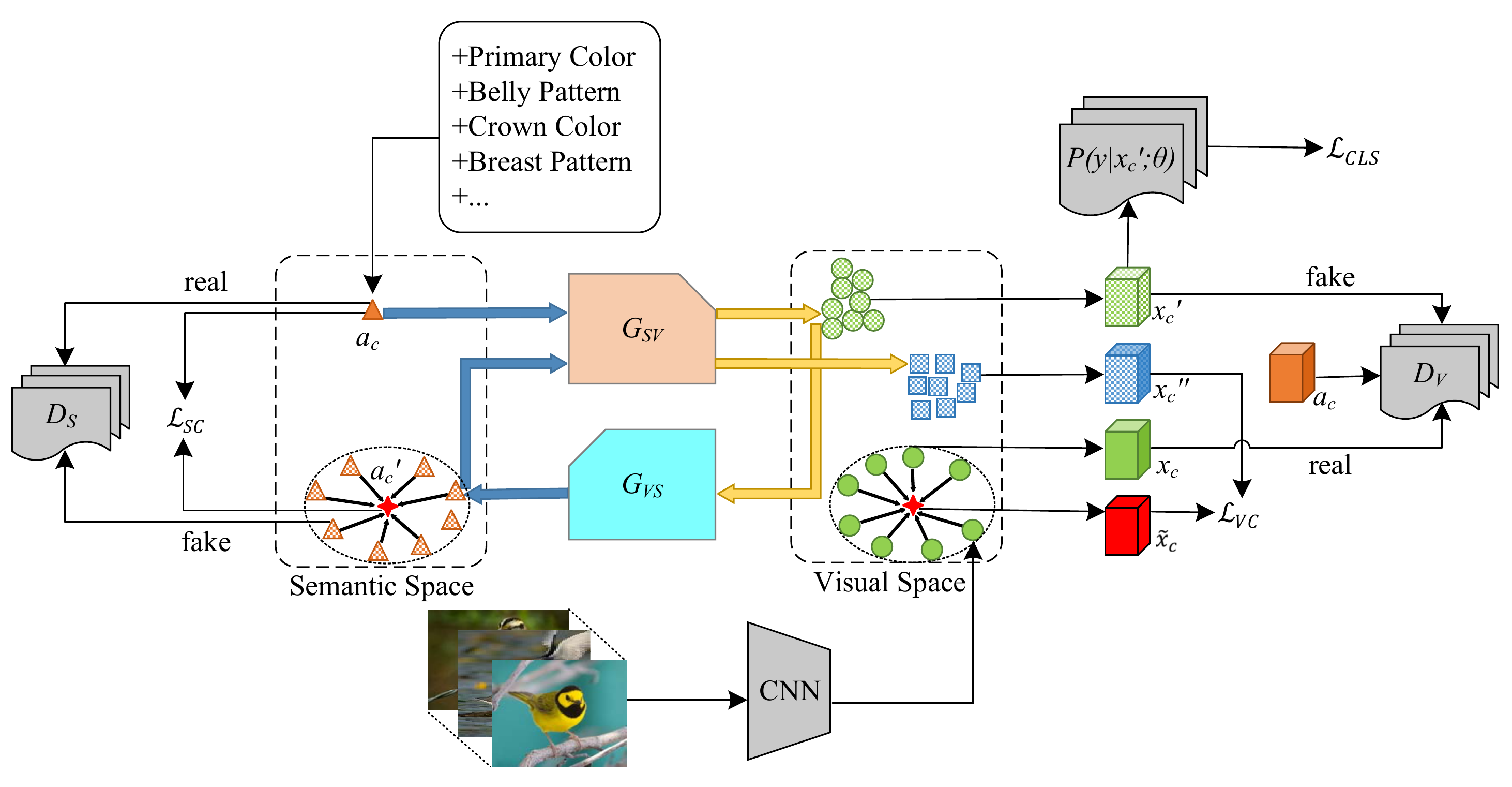}
\caption{Network architecture of DASCN. The semantic feature of class $c$, represented as $a_c$, and a group of randomly sampled noise vectors are utilized by generator $G_{SV}$ to synthesize pseudo visual features ${x_c}'$. Then the synthesized visual features are used by generator $G_{VS}$ and discriminator $D_V$ simultaneously to perform semantics-consistency constraint in the perspective of form and content and distinguish between real visual features $x_c$ and synthesized visual features ${x_c}'$. $D_S$ denotes the discriminator that distinguishes between $a_c$, and the reconstructed semantic feature ${a_c}'$ generated from corresponding ${x_c}'$. ${x_c}''$ are produced by generator $G_{SV}$ taking ${a_c}'$ and sampled noise as input to perform visual consistency constraint. Please zoom to view better.}
\label{fig_framework}
\end{figure*}

\subsection{Model Architecture}

Given the training data  $D^{Tr}$ of the seen classes, the primal task of DASCN is to learn a generator $G_{SV}$:$\mathcal{Z}\times \mathcal{A}\to \mathcal{X}$ that takes the random Gaussian noise $z\in \mathcal{Z}$ and semantic attribute $a\in \mathcal{A}$ as input to generate the visual feature  $x'\in \mathcal{X}$, while the dual task is to train an inverse generator $G_{VS}$:$\mathcal{X}\to \mathcal{A}$. Once the generator $G_{SV}$ learns to generate visual features of the seen classes conditioned on the seen class-level attributes, it can also generate that of the unseen classes. To realize this, we employ two WGANs, the primal GAN and the dual GAN. The primal GAN consists of the generator $G_{SV}$ and the discriminator $D_V$ that discriminates between fake visual features generated by $G_{SV}$ and real visual features. Similarly, the dual GAN learns a generator $G_{VS}$ and a discriminator $D_S$ that distinguishes the fake semantic features generated by $G_{VS}$ from the real data.

The overall architecture and data flow are illustrated in Figure 2. In the primal GAN, we hallucinate pseudo visual features $x_c'=G_{SV}(a_c,z)$ of the class $c$ using $G_{SV}$ based on corresponding class semantic features $a_c$ and then put the real visual features and synthetic features from $G_{SV}$ into $D_V$ to be evaluated. To ensure that $G_{SV}$ generates inter-class discrimination visual features, inspired by that work \cite{xian2018feature}, we design a classifier trained on the real visual features and minimize the classification loss over the generated features. It is formulated as:
\begin{equation}
\mathcal{L}_{CLS}=-E_{x'\sim Px'}[logP(y|x';\theta)]
\end{equation}
where $x'$ represents the generated visual feature, $y$ is the class label of $x'$, the conditional probability $P(y|x';\theta)$ is computed by a linear softmax classifier parameterized by $\theta$.

Following that is one of the main innovations in our work that we guarantee semantics-consistency in both form and content perspectives thanks to dual structure. In form, $G_{SV}(a,z)$ are translated back to semantic space using $G_{VS}$, which outputs $a'=G_{VS}\big(G_{SV}(a,z)\big)$ as the reconstruction of $a$. To achieve the goal that the generated semantic features of each class are distributed around the corresponding true semantic representation, we design the centroid regularization that regularizes the mean of generated semantic features of each class to approach respectively real semantic embeddings so as to maintain semantics-consistency to a large extent. The regularization is formulated as:
\begin{equation}
\mathcal{L}_{SC}=\frac{1}{C}\sum_{C=1}^C\Big\rVert E_{{a_c}'\sim P_{a'}^c}[{a_c}']-a_c\Big\rVert _2 
\end{equation}
where $C$ is the number of seen classes, $a_c$ is the semantic feature of class $c$, $P_{a'}^c$ denotes the conditional distribution of generated semantic features of class $c$, ${a_c}'$ are the generated features of class $c$ and the centroid is formulated as:
\begin{equation}
E_{{a_c}'\sim P_{a'}^c}[{a_c}']=\frac{1}{N_c^s}\sum_{i=1}^{N_c^s}G_{VS}\big(G_{SV}(a_c,z_i)\big)
\end{equation}
where $N_c^s$ is the number of generated semantic features of class $c$. We employ the centroid regularization to encourage $G_{VS}$ to reconstruct semantic features of each seen class that statistically match real features of that class. From the content point of view, the question of how well the pseudo semantic features $a'$ are reconstructed can be translated into the evaluation of the visual features obtained by $G_{SV}$ taking $a'$ as input. Motivated by the observation that visual features have a higher inter-class similarity and relatively lower inter-class similarity, we introduce the visual consistency constraint:
\begin{equation}
L_{VC}=\frac{1}{C}\sum_{c=1}^C\Big\rVert E_{{x_c}''\sim P_{x''}^c}\big[{x_c}''\big]-E_{{x_c}
\sim P_x^c}\big[x_c\big]\Big\rVert_2
\end{equation}
where $x_c$ is the visual features of class $c$, ${x_c}''$ is the pseudo visual feature generated by generator $G_{SV}$ employing $G_{VS}\big(G_{SV}(a_c,z)\big)$ as input, $P_x^c$ and $P_{x''}^c$ are conditional distributions of real and synthetic features respectively and the centroid of ${x_c}''$ is formulated as:
\begin{equation}
E_{{x_c}''\sim P_{x''}^c}\big[{x_c}''\big]=\frac{1}{N_c^s}\sum_{i=1}^{N_c^s}G_{SV}\Big(G_{VS}\big(G_{SV}(a_c,z_i)\big),{z_i}'\Big)
\end{equation}
It is worth nothing that our model is constrained in terms of both form and content aspects to achieve the goal of retaining semantics-consistency and achieves superior results in extensive experiments.

\subsection{Objective}

Given the issue that the Jenson-Shannon divergence optimized by the traditional GAN leads to instability training, our model is based on two WGANs that leverage the Wasserstein distance between two distributions as the objectives. The corresponding loss functions used in the primal GAN are defined as follows. First,
\begin{equation}
L_{D_V}\!=\!E_{x'\sim Px'}\big[D_V(x',a)\big]\!-\!E_{x\sim P_{data}}\big[D_V(x,a)\big]\!+\!\lambda_1E_{\hat{x}\sim P\hat{x}}\Big[\Big(\big\rVert\triangledown_{\hat{x}}D_V(\hat{x},a)\big\rVert_2-1\Big)^2\Big]
\end{equation}
where $\hat{x}=\alpha x+(1-\alpha)x'$ with $\alpha\sim U(0,1)$, $\lambda_1$ is the penalty coefficient, the first two terms approximate Wasserstein distance of the distributions of fake features and real features, the third term is the gradient penalty. Second, the loss function of the generator of the primal GAN is formulated as:
\begin{equation}
L_{G_{SV}}=-E_{x'\sim P{x'}}\big[D_V(x',a)\big]-E_{a'\sim P{a'}}\big[D_V(x,a')\big]+\lambda_2L_{CLS}+\lambda_3L_{VC}
\end{equation}
where the first two terms are Wasserstein loss, the third term is the classification loss corresponding to class labels, the forth term is visual consistency constraint introduced before, and $\lambda_1$, $\lambda_2$, $\lambda_3$ are hyper-parameters.

Similarly, the loss functions of the dual GAN are formulated as:
\begin{equation}
L_{D_S}=E_{a'\sim P{a'}}\big[D_S(a')\big]-E_{a\sim Pa}\big[D_S(a)\big]+\lambda_4E_{\hat{y}\sim P\hat{y}}\Big[\Big(\big\rVert\triangledown_{\hat{y}}D_S(\hat{y})\big\rVert_2-1\Big)^2\Big]
\end{equation}
\begin{equation}
L_{G_{VS}}=-E_{a'\sim P{a'}}\big[D_S(a')\big]+\lambda_5L_{SC}+\lambda_6L_{VC}
\end{equation}
In Eq. (8) and Eq. (9), $\hat{y}=\beta a+(1-\beta)a'$ is the linear interpolation of the real semantic feature $a$ and the fake $a'$, and $\lambda_4$, $\lambda_5$, $\lambda_6$ are hyper-parameters weighting the constraints.

\subsection{Training Procedure}

We train the discriminators to judge features as real or fake and optimize the generators to fool the discriminator. To optimize the DASCN model, we follow the training procedure proposed in WGAN \cite{gulrajani2017improved}. The training procedure of our framework is summarized in Algorithm 1. In each iteration, the discriminators $D_V$, $D_S$ are optimized for $n_1$, $n_2$ steps using the loss introduced in Eq. (6) and Eq. (8) respectively, and then one step on generators with Eq. (7) and Eq. (9) after the discriminators have been trained. According to \cite{yi2017dualgan}, such a procedure enables the discriminators to provide more reliable gradient information. The training for traditional GANs suffers from the issue that the sigmoid cross-entropy is locally saturated as discriminator improves, which may lead to vanishing gradient and need to balance discriminator and generator carefully. Compared to the traditional GANs, the Wasserstein distance is differentiable almost everywhere and demonstrates its capability of extinguishing mode collapse. We put the detailed algorithm for training DASCN model in the supplemental material.

\subsection{Generalized Zero-Shot Recognition}

With the well-trained generative model, we can elegantly generate labeled exemplars of any class by employing the unstructured component $z$ resampled from random Gaussian noise and the class semantic attribute $a_c$ into the $G_{SV}$. An arbitrary number of visual features can be synthesized and those exemplars are finally used to train any off-the-shelf classification model. For simplicity, we adopt a softmax classifier. Finally, the prediction function for an input test visual feature $v$ is:
\begin{equation}
f(v)=arg\mathop{max}\limits_{y\in\mathcal{\tilde{Y}}}P(y|v';\theta')
\end{equation}
where $\mathcal{\tilde{Y}}=\mathcal{Y}^s\cup \mathcal{Y}^u$ for GZSL.

\section{Experiments}

\subsection{Datasets Settings and Evaluation Metrics}

To test the effectiveness of the proposed model for GZSL, we conduct extensive evaluations on four benchmark datasets: CUB \cite{welinder2010caltech}, SUN \cite{patterson2012sun}, AWA1 \cite{lampert2009learning}, aPY \cite{farhadi2009describing} and compare the results with state-of-the-art approaches. Statistics of the datasets are presented in Table 1. For all datasets, we extract 2048 dimensional visual features via the 101-layered ResNet from the entire images, which is the same as \cite{xian2018feature}. For fair comparison, we follow the training/validation/testing split as described in \cite{xian2018zero}.   

\begin{table}
\centering
\caption{Datasets used in our experiments, and their statistics}
\begin{tabular}{ccccc}
\toprule
Dataset&Semantics/Dim&\# Image& \# Seen Classes&\# Unseen Classes\\
\midrule
CUB&A/312&11788&150&50\\
SUN&A/102&14340&645&72\\
AWA1&A/85&30475&40&10\\
aPY&A/64&15339&20&12\\
\bottomrule
\end{tabular}
\end{table}

At test time, in the GZSL setting, the search space includes both the seen and unseen classes, i.e. $\mathcal{Y}^u\cup \mathcal{Y}^s$. To evaluate the GZSL performance over all classes, the following measures are applied. (1) ts: average per-class classification accuracy on test images from the unseen classes with the prediction label set being  $\mathcal{Y}^u\cup \mathcal{Y}^s$. (2) tr: average per-class classification accuracy on test images from the seen classes with the prediction label set being  $\mathcal{Y}^u\cup \mathcal{Y}^s$. (3) H: the harmonic mean of above defined tr and ts, which is formulated as $H=(2\times ts\times tr)/(ts+tr)$ and quantities the aggregate performance across both seen and unseen test classes. We hope that our model is of high accuracy on both seen and unseen classes.

\subsection{Implementation Details}

Our implementation is based on PyTorch. DASCN consists of two generators and two discriminators: $G_{SV}$, $G_{VS}$, $D_V$, $D_S$. We train specific models with appropriate hyper-parameters. Due to the space limitation, here we take CUB as an example. Both the generators and discriminators are MLP with LeakyReLU activation. In the primal GAN, $G_{SV}$ has a single hidden layer containing 4096 nodes and an output layer that has a ReLU activation with 2048 nodes, while the discriminator  $D_V$ contains a single hidden layer with 4096 nodes and an output layer without activation. $G_{VS}$ and $D_S$ in the dual GAN have similar architecture with $G_{SV}$ and $D_V$ respectively. We use $\lambda_1=\lambda_4=10$ as suggested in \cite{gulrajani2017improved}. For loss term contributions, we cross-validate and set $\lambda_2=\lambda_3=\lambda_6=0.01$, $\lambda_5=0.1$. We choose noise $z$ with the same dimensionality as the class embedding. Our model is optimized by Adam with a base learning rate of $1e^{-4}$.

\subsection{Compared Methods and Experimental Results}

We compare DASCN with state-of-the-art GZSL models. These approaches fall into two categories. (1) Embedding-based approaches: CMT \cite{socher2013zero}, DEVISE \cite{frome2013devise}, ESZSL \cite{romera2015embarrassingly}, SJE \cite{akata2015evaluation}, SAE \cite{kodirov2017semantic}, LESAE \cite{liu2018zero}, SP-AEN \cite{chen2018zero}, RN \cite{sung2018learning}, KERNEL \cite{zhang2018zero}, PSR \cite{annadani2018preserving}, DCN \cite{liu2018generalized}, TRIPLE \cite{zhang2018triple}. This category suffers from the issue of the bias towards seen classes due to the lack of instances of the unseen classes. (2) Generative approaches: SE-GZSL \cite{kumar2018generalized}, GAZSL \cite{zhu2018generative}, f-CLSWGAN \cite{xian2018feature},  Cycle-WGAN \cite{felix2018multi}. This category synthesizes visual features of the seen and unseen classes and perform better for GZSL compared to the embedding-based methods.

\begin{table*}
\centering
\setlength{\tabcolsep}{1.3mm}{
\caption{Evaluations on four benchmark datasets. *indicates that Cycle-WGAN employs 1024-dim per-class sentences as class semantic rather than 312-dim per-class attributes on CUB, whose results on CUB may not be directly comparable with others.}
\begin{tabular}{l|ccc|ccc|ccc|ccc}
\toprule
&&AWA1&&&SUN&&&CUB&&&aPY\\
\midrule
Method&ts&tr&H&ts&tr&H&ts&tr&H&ts&tr&H\\
CMT \cite{socher2013zero}&0.9&87.6&1.8&8.1&21.8&11.8&7.2&49.8&12.6&1.4&\textbf{85.2}&2.8\\
DEVISE \cite{frome2013devise}&13.4&68.7&22.4&16.9&27.4&20.9&23.8&53.0&32.8&4.9&76.9&9.2\\
ESZSL	 \cite{romera2015embarrassingly}&6.6&75.6&12.1&11.0&27.9&15.8&12.6&63.8&21.0&2.4&70.1&4.6\\
SJE \cite{akata2015evaluation}&11.3&74.6&19.6&14.7&30.5&19.8&23.5&59.2&33.6&3.7&55.7&6.9\\
SAE \cite{kodirov2017semantic}&1.8&77.1&3.5&8.8&18.0&11.8&7.8&54.0&13.6&0.4&80.9&0.9\\
LESAE \cite{liu2018zero}&19.1&70.2&30.0&21.9&34.7&26.9&24.3&53.0&33.3&12.7&56.1&20.1\\
SP-AEN \cite{chen2018zero}&-&-&-&24.9&38.2&30.3&34.7&\textbf{70.6}&46.6&13.7&63.4&22.6\\
RN \cite{sung2018learning}&31.4&\textbf{91.3}&46.7&-&-&-&38.1&61.1&47.0&-&-&-\\
TRIPLE \cite{zhang2018triple}&27&67.9&38.6&22.2&38.3&28.1&26.5&62.3&37.2&-&-&-\\
f-CLSWGAN \cite{xian2018feature}&57.9&61.4&59.6&42.6&36.6&39.4&43.7&57.7&49.7&-&-&-\\
KERNEL \cite{zhang2018zero}&18.3&79.3&29.8&19.8&29.1&23.6&19.9&52.5&28.9&11.9&76.3&20.5\\
PSR \cite{annadani2018preserving}&-&-&-&20.8&37.2&26.7&24.6&54.3&33.9&13.5&51.4&21.4\\
DCN \cite{liu2018generalized}&25.5&84.2&39.1&25.5&37&30.2&28.4&60.7&38.7&14.2&75.0&23.9\\
SE-GZSL \cite{kumar2018generalized}&56.3&67.8&61.5&40.9&30.5&34.9&41.5&53.3&46.7&-&-&-\\
GAZSL	 \cite{zhu2018generative}&25.7&82.0&39.2&21.7&34.5&26.7&23.9&60.6&34.3&14.2&78.6&24.1\\
\midrule
DASCN (Ours)&\textbf{59.3}&68.0&\textbf{63.4}&42.4&\textbf{38.5}&\textbf{40.3}&\textbf{45.9}&59.0&\textbf{51.6}&\textbf{39.7}&59.5&\textbf{47.6}\\
\midrule
Cycle-WGAN* \cite{felix2018multi}&56.4&63.5&59.7&\textbf{48.3}&33.1&39.2&46.0&60.3&52.2&-&-&-\\
\bottomrule
\end{tabular}}
\end{table*}

Table 2 summarizes the performance of all the comparing methods under three evaluation metrics on the four benchmark datasets, which demonstrates that for all datasets our DASCN model significantly improves the ts measure and H measure over the state-of-the-arts. Note that Cycle-WGAN \cite{felix2018multi} employs per-class sentences as class semantic features on CUB dataset rather than per-class attributes that are commonly used by other comparison methods, so its results on CUB may not be directly comparable with others. On CUB, DASCN achieves 45.9\% in ts and 51.6\% in H, with improvements over the state-of-the-art 2.2\% and 1.9\% respectively. On SUN, it obtains 42.4\% in ts measure and 40.3\% in H measure. On AWA1, our model outperforms the runner-up by a considerable gap in H measure: 1.9\%. On aPY, DASCN significantly achieves improvements over the other best competitors 25.5\% in ts measure and 23.5\% in H measure, which is very impressive. The performance boost is attributed to the effectiveness of DASCN that imitate discriminative visual features of the unseen classes. In conclusion, our model DASCN achieves a great balance between seen and unseen classes classification and consistently outperforms the current state-of-the-art methods for GZSL.

\subsection{Ablation Study}

We now conduct the ablation study to evaluate the effects of the dual structure, the semantic centroid regularization $\mathcal{L}_{SC}$, and the visual consistency constraint $\mathcal{L}_{VC}$. We take the single WGAN model f-CLSWGAN as baseline, and train three variants of our model by keeping the single dual structure or that adding the only semantic or visual constraint, denoted as Dual-WGAN, Dual-WGAN $+\mathcal{L}_{SC}$, Dual-WGAN $+\mathcal{L}_{VC}$, respectively. Table 3 shows the performance of each setting, for the sake of simplicity, we only report the results on Apy and AWA1 datasets while similar results are presented on the other two datasets. The performance of the single Dual-WGAN drops drastically by 4.9\% on aPY and 1.4\% on AWA1 in H measure, which highlights the importance of designed semantic and visual constraints to provide an explicit supervision to our model. In the case of lacking semantic or visual unidirectional constrains, on aPY, our model drops by 1.3\% and 3.6\% respectively, while on AWA1 the gap are 0.9\% and 0.7\%. In general, the three variants of our proposed model tend to offer more superior and balanced performance than the baseline while DASCN incorporates dual structure, semantic centroid regularization and visual consistency constraint into a unified framework and achieves the best improvement, which demonstrates that different components promote each other and work together to improve the performance of DASCN significantly.

\begin{table}
\centering
\caption{Effects of different components on aPY and AWA1 datasets with GZSL setting.}
\begin{tabular}{l|ccc|ccc}
\toprule
&&aPY&&&AWA1&\\
\midrule
Methods&ts&tr&H&ts&tr&H\\
WGAN-baseline&32.4&57.5&41.4&56.5&62.4&59.3\\
Dual-WGAN&34.1&57.0&42.7&57.5&67.4&62.0\\
Dual-WGAN $+\mathcal{L}_{SC}$&35.4&58.2&44.0&57.7&\textbf{68.6}&62.7\\
Dual-WGAN $+\mathcal{L}_{VC}$&36.7&\textbf{62.0}&46.3&58.3&67.3&62.5\\
DASCN&\textbf{39.7}&59.5&\textbf{47.6}&\textbf{59.3}&68.0&\textbf{63.4}\\
\bottomrule
\end{tabular}
\end{table} 

\begin{figure}[htbp]
\centering
\subfigure[]{
\begin{minipage}[t]{0.33 \linewidth}
\centering
\includegraphics[width=1.6 in]{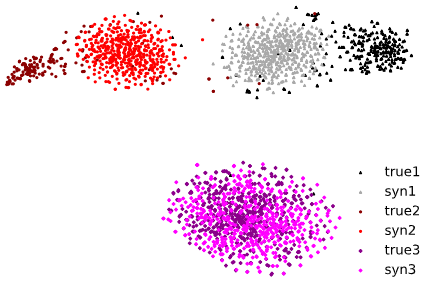}
\end{minipage}%
}%
\subfigure[]{
\begin{minipage}[t]{0.33 \linewidth}
\centering
\includegraphics[width=1.8 in]{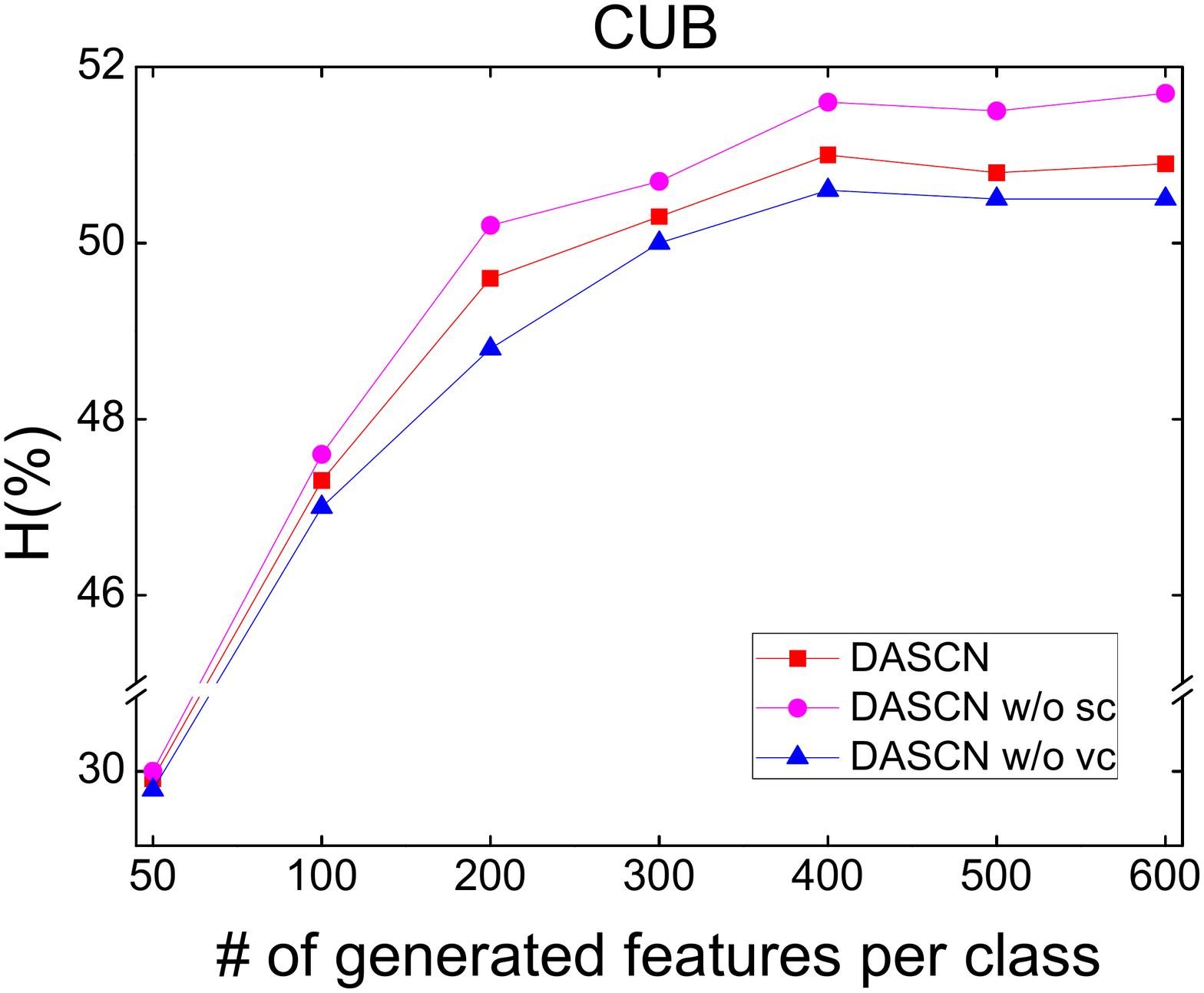}
\end{minipage}%
}%
\subfigure[]{
\begin{minipage}[t]{0.33 \linewidth}
\centering
\includegraphics[width=1.8 in]{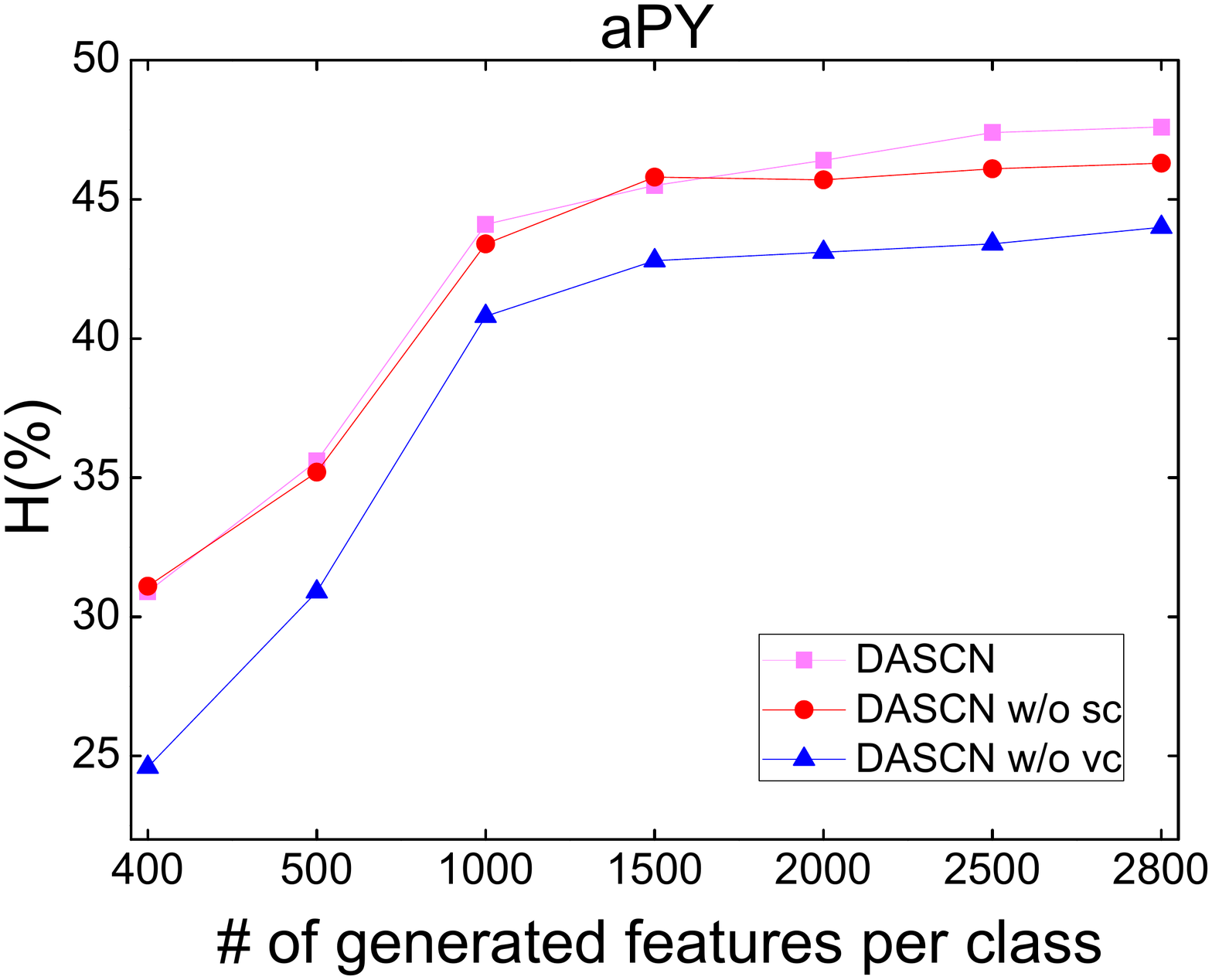}
\end{minipage}%
}%
\centering
\caption{\textbf{(a)}: t-SNE visualization of real visual feature distribution and synthesized feature distribution from randomly selected three unseen classes; \textbf{(b, c)}: Increasing the number of samples generated by DASCN and its variants wrt harmonic mean H. DASCN w/o SC denotes DASCN without semantic consistency constraint and DASCN w/o VC stands for that without visual consistency constraint.}
\end{figure}

\subsection{Quality of Synthesized Samples}

We perform an experiment to gain a further insight into the quality of the generated samples, which is one key issue of our approach, although the quantitative results reported for GZSL above demonstrate that the samples synthesized by our model are of significant effectiveness for GZSL task. Specifically, we randomly sample three unseen categories from aPY and visualize both true visual features and synthesized visual features using t-SNE \cite{maaten2008visualizing}. In Figure 3a, the empirical distributions of generated and true visual features are of intra-class diversity and inter-class separability respectively, which intuitively demonstrates that not only the synthesized feature distributions well approximate the true distributions but also our model introduces high discriminative power of the synthesized features to a large extent.

Finally, we evaluate how the number of the generated samples per class affects the performance of DASCN and its variants. Obviously, as shown in Figure 3b, 3c, we notice not only that H increases with an increasing number of synthesized samples and asymptotes gently, but also that DASCN with visual-semantic interactions achieves better performance in all circumstance, which further validates the superiority and rationality of different components of our model.

\section{Conclusion}

We propose a novel generative model for GZSL to synthesize inter-class discrimination and semantics-preserving visual features for seen and unseen classes. The DASCN architecture consists of the primal GAN and the dual GAN to collaboratively promote each other, which captures the underlying data structures of visual and semantic representations and enhances the knowledge transfer to unseen categories, as well as alleviates the inherent semantic loss problem for GZSL. We conduct extensive experiments on four benchmark datasets. Extensive experimental results consistently demonstrates the superiority of DASCN to state-of-the-art GZSL approaches.

\bibliographystyle{unsrt}

\end{document}